\documentclass[conference]{IEEEtran}
\IEEEoverridecommandlockouts

\usepackage{cite}
\usepackage{amsmath,amssymb,amsfonts}
\usepackage{graphicx}
\usepackage{stfloats}
\usepackage{textcomp}
\usepackage{xcolor}
\usepackage{hyperref}
\usepackage{colortbl}

\usepackage{algorithm}
\usepackage{algpseudocode}

\usepackage{subcaption}

\usepackage[none]{hyphenat}

\usepackage{times}         
\usepackage{latexsym}      
\usepackage[T1]{fontenc}   
\usepackage[utf8]{inputenc}
\usepackage{microtype}     
\usepackage{inconsolata}   
\usepackage{multicol,multirow} 
\usepackage{enumitem}      
\usepackage{booktabs}      
\usepackage{tabularx}      
\usepackage{makecell}      
\usepackage{array}         
\usepackage{bbding}        
\usepackage{color}         

\def\BibTeX{{\rm B\kern-.05em{\sc i\kern-.025em b}\kern-.08em
    T\kern-.1667em\lower.7ex\hbox{E}\kern-.125emX}}
\begin{document}

\makeatletter
\newcommand{\linebreakand}{%
  \end{@IEEEauthorhalign}
  \hfill\mbox{}\par
  \mbox{}\hfill\begin{@IEEEauthorhalign}
}
\makeatother

\newcommand{\wxb}[1]{\textcolor{blue}{[#1]}}

\title{
    Aspect-Based Summarization with Self-Aspect Retrieval Enhanced Generation
\\
}
\author{
    \IEEEauthorblockN{\small  Yichao Feng}
    \IEEEauthorblockA{\small
        \textit{College of Computing and Data Science} \\
        \textit{Nanyang Technological University} \\
        Singapore, Singapore \\
        yichao.feng@ntu.edu.sg
    }
    \and
    \IEEEauthorblockN{\small  Shuai Zhao}
    \IEEEauthorblockA{\small
        \textit{College of Computing and Data Science} \\
        \textit{Nanyang Technological University} \\
        Singapore, Singapore \\
        shuai.zhao@ntu.edu.sg
    }
    \and
    \IEEEauthorblockN{\small  Yueqiu Li}
    \IEEEauthorblockA{\small
        \textit{School of Humanities} \\
        \textit{Nanyang Technological University} \\
        Singapore, Singapore \\
        LIYU0069@e.ntu.edu.sg
    }
    \linebreakand
    \IEEEauthorblockN{\small  Luwei Xiao}
    \IEEEauthorblockA{\small
        \textit{School of Computer Science and Technology} \\
        \textit{East China Normal University} \\
        Shanghai, China \\
        louisshaw@stu.ecnu.edu.cn
    }
    \and
    \IEEEauthorblockN{\small  Xiaobao Wu}
    \IEEEauthorblockA{\small
        \textit{College of Computing and Data Science} \\
        \textit{Nanyang Technological University} \\
        Singapore, Singapore \\
        xiaobao002@e.ntu.edu.sg
    }
    \and
    \IEEEauthorblockN{\small  Anh Tuan Luu}
    \IEEEauthorblockA{\small
        \textit{College of Computing and Data Science} \\
        \textit{Nanyang Technological University} \\
        Singapore, Singapore \\
        anhtuan.luu@ntu.edu.sg
    }
}

\maketitle

\begin{abstract}
Aspect-based summarization aims to generate summaries tailored to specific aspects, addressing the resource constraints and limited generalizability of traditional summarization approaches. Recently, large language models have shown promise in this task without the need for training. However, they rely excessively on prompt engineering and face token limits and hallucination challenges, especially with in-context learning. To address these challenges, in this paper, we propose a novel framework for aspect-based summarization: Self-Aspect Retrieval Enhanced Summary Generation. Rather than relying solely on in-context learning, given an aspect, we employ an embedding-driven retrieval mechanism to identify its relevant text segments. This approach extracts the pertinent content while avoiding unnecessary details, thereby mitigating the challenge of token limits. Moreover, our framework optimizes token usage by deleting unrelated parts of the text and ensuring that the model generates output strictly based on the given aspect. With extensive experiments on benchmark datasets, we demonstrate that our framework not only achieves superior performance but also effectively mitigates the token limitation problem.

\end{abstract}


\section{Introduction}

Aspect-based summarization (ABS) produces concise summaries tailored to specific user needs, addressing the limitations of general summarization by focusing on targeted, nuanced information. This approach has proven particularly effective for specialized domains, such as customer sentiment analysis \cite{zhang2024senticvec, xiao2024vanessa}, financial analysis \cite{mao2023discovering, du2024evaluation}, legislative proposals \cite{gesnouin2024Llamandement}, and legal documents \cite{jain2024sentence}. ABS integrates seamlessly with large language models (LLMs), which have revolutionized summarization by offering significant advantages over traditional methods \cite{mao2022biases,wu2020short,wu2022mitigating,wu2024fastopic,wu2024survey}.

Research shows that LLM-generated summaries are often preferred by humans over those produced by earlier techniques \cite{zhang2024benchmarking}. The integration of ABS with LLMs offers new opportunities for improving summarization, enabling the creation of summaries that are both contextually relevant and application-specific. This synergy has the potential to improve summarization performance in real-world applications and address the growing demand for aspect-oriented, domain-specific summarization.

However, despite their advantages, the existing LLM-based summarization approaches discussed above face several critical challenges. First, most LLMs are constrained by limited input context length \cite{xue2024repeat}, making it difficult to effectively summarize lengthy texts. While some models \cite{chen2023longlora, zhang2024bench} have been developed to handle extended inputs, they often struggle to produce accurate summaries due to the increased complexity. This constraint forces LLMs to truncate or omit critical details, especially when relying on techniques like in-context learning (ICL). Furthermore, ICL, which involves providing relevant examples or prompts within the input, reduces the available token space for the actual content, further complicating the generation of precise, aspect-focused summaries. Second, LLMs often lack the ability to apply selective attention, processing the entire input indiscriminately rather than focusing on the most relevant parts. This lack of focus exacerbates the problem of hallucination—generating content that is either irrelevant or factually incorrect—particularly when dealing with long or complex texts \cite{huang2023survey}. Unlike humans, who can synthesize information from extended inputs while maintaining accuracy, LLMs may introduce inconsistencies or fabricate details, leading to degraded performance in aspect-based summarization. Addressing these challenges is crucial to improving the reliability and utility of LLM-generated summaries.

To address these challenges, we propose a novel retrieve-and-prune method called Self-Aspect Retrieval Enhanced Summary Generation (SARESG). Instead of feeding the entire text or relying solely on prompting methods, we perform a dense retrieval task using an embedding model. This retrieval process extracts the most relevant chunks of text, focusing exclusively on sections directly related to the desired aspect of the summary. The retrieval process is recursive, continuing until the text is pruned to the desired length. Pruning not only helps mitigate the issue of token limits but also addresses the challenge of hallucination. By eliminating irrelevant and redundant content, the model's attention is directed exclusively toward pertinent information, reducing the likelihood of generating fabricated or inaccurate details. Shorter, aspect-focused inputs help the model maintain factual consistency and produce summaries that are highly aligned with the specified aspect. In addition, this pruning process simplifies input, making it easier for the model to synthesize and process the information effectively. To further enhance the summarization process, a re-ranking mechanism is employed after the retrieval and pruning steps, helping the model better understand the context of the remaining text. By preserving valuable token space, our method allows for the integration of ICL techniques \cite{dong-etal-2024-survey}, such as guiding prompts or system instructions, which further refine the generation process. Ultimately, by efficiently focusing on relevant content, optimizing token usage, and reducing the risk of hallucination, our method significantly improves the precision, relevance, and reliability of LLM-generated summaries, unlocking the full potential of LLMs for ABS tasks.

We conducted comprehensive experiments to evaluate the robustness and efficiency of our system. The experimental results demonstrate that our approach consistently generates more accurate, aspect-aligned summaries across diverse datasets. Moreover, it enables more efficient text pruning compared to traditional truncation methods commonly used in ABS systems, while achieving superior performance through the integration of ICL techniques.

The main contributions of this paper can be concluded as follows:
\begin{itemize}[leftmargin=*]
    \item
        We proposed a self-retrieval mechanism that effectively eliminates text segments irrelevant to the desired aspect of the summary. This approach enhances the precision of ABS by guiding the model's attention to the most pertinent information.
    \item
        Our method conserves a token space, enabling the use of ICL and system prompts, which improves the customization and adaptability of the summarization process.
    \item
        We conduct extensive experiments across various language models and datasets, demonstrating the strong performance of our proposed method. Detailed ablation studies further highlight the robustness of our approach and the effectiveness of incorporating ICL. 
\end{itemize}

\section{Related Work}
\paragraph{Aspect-based Summarization}
ABS aims to generate summaries that are organized around specific aspects or attributes of the input content \cite{frermann2019inducing}, rather than providing a generic overview. It has been extensively studied in both structured and unstructured data contexts, including customer reviews \cite{mao2021bridging, xiao2024atlantis}, legal decisions \cite{santosh2024lexabsumm}, and news documents \cite{ahuja2021aspectnews}. The goal is to extract or generate summaries that highlight key opinions, facts, or narratives associated with predefined or inferred aspects. This task is particularly important in applications such as product review analysis, where users may prioritize different aspects (e.g., "price," "quality," "durability"), or in domains like opinion mining on social or political topics.

Traditional ABS methods often adopt a two-step approach: first, extract relevant content and then generate a summary\cite{zhong-etal-2020-extractive}. Some studies have also explored training Longformer models \cite{beltagy2020longformer}, leveraging their extended attention capabilities to handle long sequences.

\paragraph{LLM-based Summarization}

The emergence of LLMs, such as Llama and ChatGPT, has revolutionized the field of natural language processing\cite{zhao2024feamix,zhao2024exploring,wu2024akew,wu2024antileak,pan2024fallacy, chencan}, including text summarization. These models are trained in vast corpora of text, enabling them to capture nuanced linguistic patterns, contextual dependencies, and domain-specific knowledge. Consequently, LLMs have demonstrated remarkable capabilities in both extractive and abstractive summarization tasks, significantly outperforming traditional and task-specific approaches~\cite{he2023survey, mao2025survey}.

The current work involves two types of structures. The first structure focuses on training and tuning. Some systems employ few-shot tuning methods \cite{navarro2022few}, where the model is fine-tuned on a small dataset, or parameter-efficient fine-tuning methods \cite{han2024parameter}. However, these approaches often lack generalizability and are time-consuming. The second structure involves utilizing prompting techniques. This includes crafting carefully designed prompts for specific tasks \cite{white2023prompt} or leveraging ICL for summarization tasks \cite{jain2023multi,zhao2023softmax}.

\section{Methodology}

\begin{figure*}[!t]
    \centering
    \includegraphics[width=0.9\linewidth]{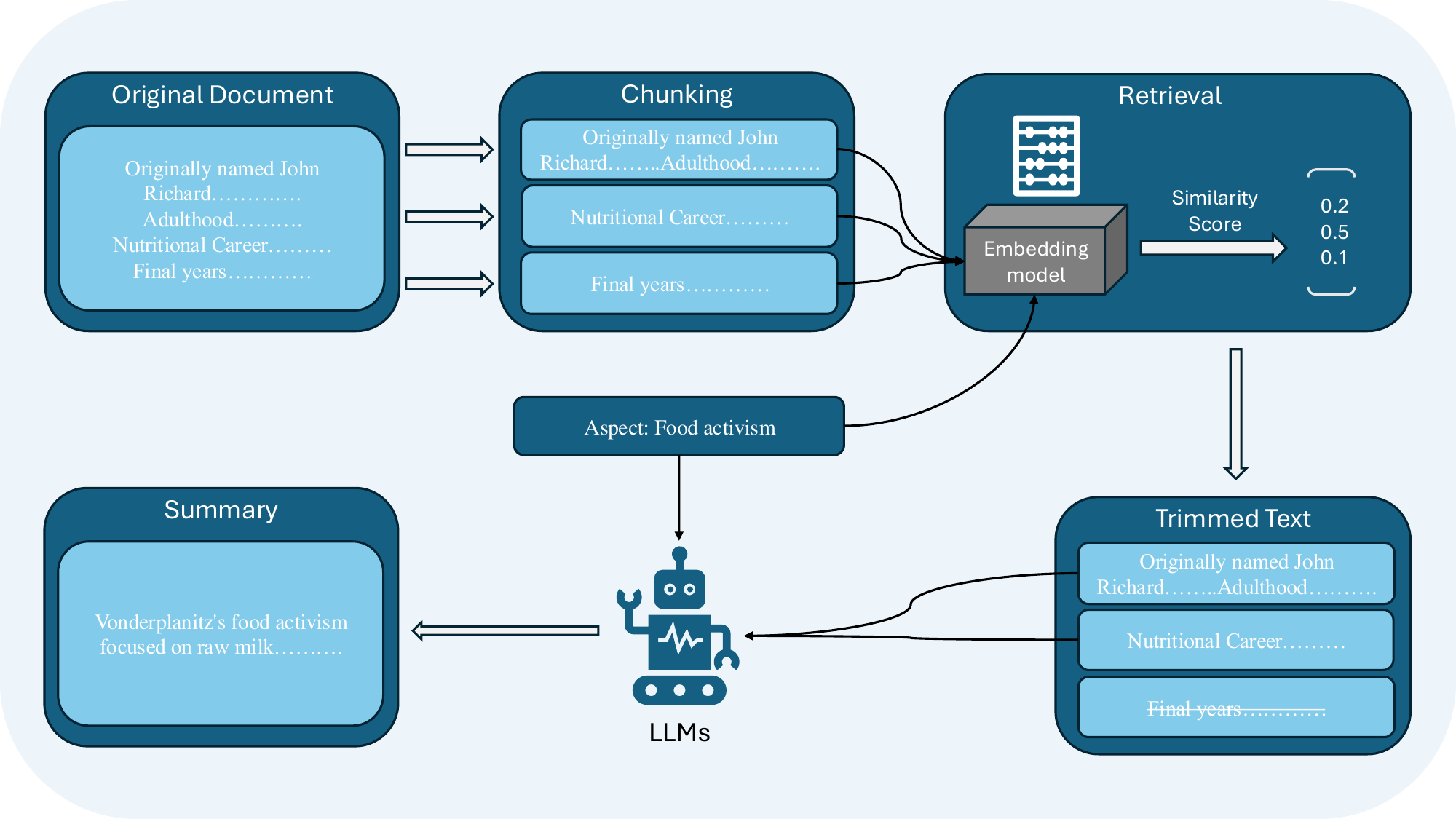}
    \caption{\textbf{This diagram illustrates the generation process of the system. The model first evaluates the overall length of the input text. If the text is shorter than a predefined threshold, it is directly fed into the LLM for summary generation. For longer texts, the system splits the input into smaller chunks, uses an embedding model to assign relevance scores to each chunk, filters out the least relevant parts, and reconstructs the remaining content in its original order to ensure readability and coherence for the model.}}
    \label{fig:main_rag_diagram}
\end{figure*}

In this section, we will introduce our SARESG system. As illustrated in Figure~\ref{fig:main_rag_diagram} which provides an overview of our approach. Where we utilized the power of embedding model to retrieve and prune the unrelated part of the document and generate an aspect based summary. By doing so, we ensure that the content fed into the summarization model is highly relevant to the specific aspects being analyzed. This approach allows for the generation of more precise and focused ABS.

Our method also supports the processing of longer input texts, a feature that enhances the richness and depth of the summaries. Additionally, by incorporating retrieval, we enable the use of ICL for the summarization task, making it applicable across a wide range of datasets. This adaptability is crucial for achieving high-quality summaries in diverse domains.

\subsection{Sentences retrieval}
We begin by splitting the document \( D \) into multiple sequential chunks, represented as
\begin{equation}
D = \{ d_{1}, d_{2}, d_{3}, \dots, d_{n} \},
\end{equation}
where each chunk \( d_{i} \) (except for the last one) contains exactly 256 words. This fixed chunk size ensures that each \( d_{i} \) contains enough contextual information to enable effective extraction of relevant content. Let \( A \) denote the desired topic or aspect that we aim to extract from the document. A further analysis was conducted in Section~\ref{subsec:influence_chunk_size_sentence_retrieval} to demonstrate that the model achieves higher performance when using chunking instead of directly processing sentences.

Next, we introduce an embedding model \( f(\cdot) \) that maps each chunk \( d_{i} \) and the aspect \( A \) to their respective embedding representations in a high-dimensional semantic space. We define the embedding of \( d_{i} \) as \( f(d_{i}) \) and the embedding of \( A \) as \( f(A) \). For each chunk \( d_{i} \), we calculate the similarity \( S_{i} \) between \( f(d_{i}) \) and \( f(A) \) using a similarity measure. Common choices include cosine similarity.

Within each chunk \( d_{i} \), let \( d_{i} = \{ s_{i,1}, s_{i,2}, \dots, s_{i,m_{i}} \} \), where \( s_{i,j} \) represents the \( j \)-th sentence in chunk \( d_{i} \) and \( m_{i} \) is the total number of sentences in \( d_{i} \). For each sentence \( s_{i,j} \), we compute the similarity score \( S_{i,j} \) with respect to \( A \):
\begin{equation}
S_{i,j} = \cos(f(s_{i,j}), f(A)) = \frac{f(s_{i,j}) \cdot f(A)}{\|f(s_{i,j})\| \|f(A)\|}.
\end{equation}

Next, we rank all sentences \( s_{i,j} \) within each chunk \( d_{i} \) by their similarity scores \( S_{i,j} \) in descending order. We then select the top sentences with the highest \( S_{i,j} \) values until the cumulative word count of the selected sentences reaches a predefined threshold \( W \) (total number of words). This subset of selected sentences from each chunk is denoted as \( d_{i}^{\text{pruned}} \), and can be formally defined as:
\begin{equation}
d_{i}^{\text{pruned}} = \text{Top-W} \{ s_{i,j} : S_{i,j} \},
\end{equation}
where \( \text{Top-W} \{ s_{i,j} : S_{i,j} \} \) returns the sentences ordered by \( S_{i,j} \) until the cumulative word count meets the word threshold \( W \) for each chunk \( d_{i} \). 

After selecting the most relevant sentences within each chunk, we rearrange them in their original sequential order as they appeared in \( D \) to maintain coherence. Finally, we define the pruned document \( D_{\text{pruned}} \) as the concatenation of all pruned chunks \( d_{i}^{\text{pruned}} \) in their original order:
\begin{equation}
D_{\text{pruned}} = \{ d_{1}^{\text{pruned}}, d_{2}^{\text{pruned}}, \dots, d_{n}^{\text{pruned}} \}.
\end{equation}

The final pruned document \( D_{\text{pruned}} \) preserves the original sequential order of sentences that have the highest relevance to the aspect \( A \), while adhering to the predefined word limit, thereby forming a condensed, aspect-specific representation of the original document \( D \).

\section{Experiments}
\newcommand{\pub}[1]{\color{red}{\tiny{#1}}}
\newcommand{\Frst}[1]{{\textbf{#1}}}
\newcommand{\Scnd}[1]{{\underline{#1}}}
\begin{table*}[!t]
\centering
\footnotesize
\renewcommand{\arraystretch}{1.35}
\setlength{\tabcolsep}{4.0mm}
\vspace{-1.5mm}
{
\begin{tabular}{l|p{80pt}|c|ccc|ccc}
    \toprule[1.5pt]
    \multirow{2}{*}{\textbf{\#}} & \multirow{2}{*}{\textbf{Method}} & \multirow{2}{*}{\textbf{METEOR}} 
    & \multicolumn{3}{c|}{\textbf{ROUGE}} & \multicolumn{3}{c}{\textbf{BERTSCORE}} \\ 
    \cmidrule(rl){4-6}\cmidrule(rl){7-9}
    & & & \textbf{R1} & \textbf{R2} & \textbf{RL} & \textbf{Precision} & \textbf{Recall} & \textbf{F1} \\    
    \noalign{\hrule height 1.5pt}
    \rowcolor{gray!20}\multicolumn{9}{c}{\it{\textbf{MA-news}}} \\
    \hline
    1& Original         & 22.91 & 18.21 & 5.63  & 12.38 & 53.64 & 44.04 & 48.06 \\
    2& Selective Context         & 20.45 & 17.01 & 4.24  & 11.34 & 51.33 & 43.54 & 46.87 \\
    3& \textbf{SARESG}                & \textbf{24.07} & \textbf{19.62} & \textbf{6.20} & \textbf{13.28} & \textbf{54.77} & \textbf{45.22} & \textbf{49.24} \\
    \hline
    5& Truncated\_ICL    & 25.33 & 23.66 & 7.28  & 15.24 & \textbf{62.83} & 46.88 & \textbf{53.24} \\
    6& \textbf{SARESG\_ICL}          & \textbf{26.83} & \textbf{24.20} & \textbf{7.31} & \textbf{15.51} & 58.02 & \textbf{48.97} & 52.91 \\
    \hline
    \rowcolor{gray!20}\multicolumn{9}{c}{\it{\textbf{OAsum}}} \\
    \hline
    1& Original     & 22.39 & 16.51 & 5.80   & 11.37 & 60.73 & 45.17 & 51.28 \\
    2& Selective Context   & 20.47 & 14.39 & 4.59  & 9.73  & 59.32 & 43.10 & 49.44 \\
    3& \textbf{SARESG}                & \textbf{23.48} & \textbf{18.04} & \textbf{6.23} & \textbf{12.50} & \textbf{60.94} & \textbf{45.99} & \textbf{51.87} \\
    \hline
    5& Truncated\_ICL    & 25.33 & 19.06 & \textbf{6.30} & 12.70  & \textbf{62.83} & 46.88 & 53.24 \\
    6& \textbf{SARESG\_ICL}   & \textbf{25.56} & \textbf{19.72} & 6.22  & \textbf{13.08} & 62.53 & \textbf{47.45} & \textbf{53.46} \\
    \hline
    \rowcolor{gray!20}\multicolumn{9}{c}{\it{\textbf{USB}}} \\
    \hline
    1& Original     & 20.56 & 12.56 & 4.58  & 9.07  & 62.37 & 41.82 & 49.62 \\
    2& Selective Context   & 21.42 & 13.63 & 4.28  & 9.58  & 61.33 & 42.64 & 49.88 \\
    3& \textbf{SARESG}             & \textbf{23.69} & \textbf{16.36} & \textbf{5.85} & \textbf{11.78} & \textbf{62.58} & \textbf{44.45} & \textbf{51.41} \\
    \hline
    5& Truncated\_ICL       & 26.47 & 17.60  & 6.22  & 12.47 & 64.32 & 46.01 & 53.15 \\
    6& \textbf{SARESG\_ICL}         & \textbf{27.01} & \textbf{18.54} & \textbf{6.59} & \textbf{12.94} & \textbf{64.59} & \textbf{46.79} & \textbf{53.75} \\
 \bottomrule[1.5pt]
\end{tabular}
}
\caption{\textbf{The model's performance with Mistral 8$\times$7b was evaluated across three datasets using three major types of metrics. Methods 1, 2, and 3 represent the results of direct zero-shot inference, while methods 5 and 6 were obtained through ICL with one sample.}}
\label{tab:mistral8*7b}
\end{table*}
\subsection{Experiment Setup}

In this section, we introduce the experimental details, which include the datasets, metrics, models, and experimental detail.


\noindent{\bf Datasets} To evaluate the robustness of our method, we conducted experiments on three ABS datasets: USB\cite{krishna2023usb}, OAsum\cite{yang2023OAsum}, and Ma-news\cite{frermann-klementiev-2019-inducing}. Each dataset differs in average text length and number of aspects, allowing us to test our approach across varied data characteristics. Due to the constrain of the computational resources, we selected the first 2000 rows of data for Ma-news and OAsum. Furthermore, due to the token limit in the ICL setting, overly long demonstration examples may lead to token depletion. Therefore, we only retain samples with a length of less than 1024.

\noindent{\bf Metrics }We evaluated the generated summaries using several widely-adopted metrics, including METEOR\cite{banerjee2005meteor}, ROUGE (ROUGE-1, ROUGE-2, and ROUGE-L)\cite{lin-2004-rouge}, Precision, Recall, and F1 score with BERTScore\cite{zhang2019bertscore}.

\noindent{\bf Experiment Details}
To evaluate our approach, we experimented with two families of LLMs: Llama3 \cite{dubey2024Llama} and Mistral 8$\times$7b \cite{jiang2024mixtral}, which have distinct architectures. To investigate the impact of model scale, we included two versions of Llama3: Llama3-8b and Llama3-70b. This allowed us to explore how model size influences performance, particularly in summarization tasks, with a focus on coherence and capturing nuanced aspects of the text. And the retrieval model we used were the jasper model on top of the stella A model called \href{https://huggingface.co/dunzhang/stella_en_1.5B_v5/tree/main}{stella en 1.5b}.

For evaluation, we conducted zero-shot and one-shot experiments. These experiments included both standard and selective context-based methods\cite{li-etal-2023-compressing} to reduce the length of text to similar length.
to measure the trade-offs between computational efficiency and model performance.

\paragraph{Zero-shot Generation}
In the zero-shot setting, we evaluated three methods. The first, serving as the baseline, used the full original text as input, allowing the model to independently generate ABS. The second method, based on the Selective Context approach \cite{li-etal-2023-compressing}, identified and pruned redundant portions of the input text, feeding only the most relevant segments into the model to generate summaries. The third method, our own adaptation of Selective Context, utilized an embedding model to identify text chunks related to specific aspects. These pruned chunks were used for ABS. For a fair comparison, the input length for both the second and third methods was adjusted to ensure consistency with the baseline.

\paragraph{One-shot Generation with Selective Context}
In the one-shot setting, we included a guiding example within the input prompt. To address the token limit of 4096 tokens, we applied a truncation strategy inspired by prior work \cite{ravaut2024context, yang2023exploring, li2023learning}. Starting with the document, we incrementally truncated words from the end until the input fit within the token limit. For longer datasets, such as USB and OAsum, we ensured fairness by selecting the guiding example from the top 20 shortest documents in the training set. This method preserved the contextual utility of the guide while balancing input length constraints. 

\section{Results}
\begin{table*}[!t]
\centering
\footnotesize
\renewcommand{\arraystretch}{1.2}  
\setlength{\tabcolsep}{3.0mm}       
\vspace{-1.5mm}
{
\begin{tabular}{l|p{80pt}|c|ccc|ccc}
    \toprule[1.5pt]
    \multirow{2}{*}{\textbf{\#}} & \multirow{2}{*}{\textbf{Method}} & \multirow{2}{*}{\textbf{METEOR}} 
    & \multicolumn{3}{c|}{\textbf{ROUGE}} & \multicolumn{3}{c}{\textbf{BERTSCORE}} \\ 
    \cmidrule(rl){4-6}\cmidrule(rl){7-9}
    & & & \textbf{R1} & \textbf{R2} & \textbf{RL} & \textbf{Precision} & \textbf{Recall} & \textbf{F1} \\    
    \noalign{\hrule height 1.5pt}
    \rowcolor{gray!20}\multicolumn{9}{c}{\it{\textbf{MA-news}}} \\
    \hline
    1& Original         & 15.86 & 10.81 & 3.44  & 7.60  & 56.36 & 36.19 & 43.70 \\
    2& Selective Context         & 15.42 & 7.99  & 2.09  & 5.39  & 55.73 & 37.70 & 44.78 \\
    3& \textbf{SARESG}                & \textbf{17.10} & \textbf{11.22} & \textbf{3.69} & \textbf{7.88} & \textbf{57.55} & \textbf{37.66} & \textbf{45.16} \\
    \hline
    5& Truncated\_ICL    & \textbf{28.48} & 25.93 & 8.36  & 16.81 & \textbf{59.47} & 50.40 & 54.40 \\
    6& \textbf{SARESG\_ICL}          & 27.86 & \textbf{27.05} & \textbf{8.52} & \textbf{17.48} & 58.51 & \textbf{51.38} & \textbf{54.54} \\
    \hline
    \rowcolor{gray!20}\multicolumn{9}{c}{\it{\textbf{OAsum}}} \\
    \hline
    1& Original    & 13.27 & 7.17  & 2.23  & 5.20  & 57.03 & 38.86 & 45.74 \\
    2& Selective Context   & 13.62 & 7.51  & 2.06  & 5.43  & 57.69 & 39.08 & 46.22 \\
    3& \textbf{SARESG}                & \textbf{15.98} & \textbf{9.23} & \textbf{3.04} & \textbf{6.56} & \textbf{60.22} & \textbf{39.45} & \textbf{47.17} \\
    \hline
    5& Truncated\_ICL    & \textbf{24.83} & 21.43 & \textbf{6.86} & 14.44 & \textbf{59.53} & 46.86 & 52.01 \\
    6& \textbf{SARESG\_ICL}   & 24.82 & \textbf{22.28} & 6.85  & \textbf{14.96} & 59.02 & \textbf{47.74} & \textbf{52.35} \\
    \hline
    \rowcolor{gray!20}\multicolumn{9}{c}{\it{\textbf{USB}}} \\
    \hline
    1& Original     & 11.06 & 5.89  & 2.02  & 4.45  & 58.11 & 34.64 & 42.90 \\
    2& Selective Context   & 13.40 & 6.89  & 1.74  & 5.16  & 57.41 & 37.14 & 44.74 \\
    3& \textbf{SARESG}             & \textbf{16.00} & \textbf{8.16} & \textbf{2.97} & \textbf{6.17} & \textbf{61.62} & \textbf{37.33} & \textbf{46.09} \\
    \hline
    5& Truncated\_ICL       & 29.74 & 23.70 & 9.51  & 17.22 & 63.28 & 49.29 & 54.86 \\
    6& \textbf{SARESG\_ICL}         & \textbf{31.17} & \textbf{25.59} & \textbf{9.98} & \textbf{18.16} & \textbf{64.30} & \textbf{50.81} & \textbf{56.25} \\
 \bottomrule[1.5pt]
\end{tabular}
}
\caption{\textbf{The model's performance with Llama3-8b across three datasets showed that our method was able to outperform most metrics. Additionally, with ICL, the 8b model could perform on par with the 70b model.}}
\label{tab:Llama8b}
\end{table*}

\subsection{Main Results}

The experimental results for the Mistral and Llama3 models highlight the consistent performance advantage of the SARESG method across multiple datasets and metrics (Table~\ref{tab:mistral8*7b}, Table~\ref{tab:Llama8b}, and Table~\ref{tab:Llama70b}).

For the Mistral model, SARESG outperformed the Original and Selective Context methods across all datasets. In the Ma-news dataset, SARESG achieved a METEOR score of 24.07, exceeding the Original’s 22.91 and Selective Context’s 20.45, with similar improvements across ROUGE and BERTScore metrics. Comparable trends were observed in the OAsum and USB datasets, where SARESG consistently delivered superior results. Additionally, SARESG\_ICL demonstrated stronger performance than Truncated, achieving the highest scores in METEOR, ROUGE, and BERTScore metrics(e.g., 27.01 METEOR on USB).

For the Llama3 models, SARESG maintained its advantage. In the Llama3-8b model, SARESG consistently outperformed baselines, achieving a METEOR score of 17.10 in the Ma-news data set versus 15.86 (original) and 15.42 (selective context). Similar improvements were observed in the OAsum and USB datasets. The trend continued with the Llama3-70b model, where SARESG excelled, particularly in the SARESG\_ICL setting, achieving METEOR scores of 30.41 (Ma-news) and 32.65 (USB), alongside superior ROUGE and BERTScore values.

These results highlight SARESG’s adaptability and effectiveness in generating high-quality, contextually relevant summaries, consistently outperforming baselines across diverse datasets and model configurations.

\begin{table*}[!t]
\centering
\footnotesize
\renewcommand{\arraystretch}{1.2}  
\setlength{\tabcolsep}{3.0mm}       
\vspace{-1.5mm}
{
\begin{tabular}{l|p{80pt}|c|ccc|ccc}
    \toprule[1.5pt]
    \multirow{2}{*}{\textbf{\#}} & \multirow{2}{*}{\textbf{Method}} & \multirow{2}{*}{\textbf{METEOR}} 
    & \multicolumn{3}{c|}{\textbf{ROUGE}} & \multicolumn{3}{c}{\textbf{BERTSCORE}} \\ 
    \cmidrule(rl){4-6}\cmidrule(rl){7-9}
    & & & \textbf{R1} & \textbf{R2} & \textbf{RL} & \textbf{Precision} & \textbf{Recall} & \textbf{F1} \\    
    \noalign{\hrule height 1.5pt}
    \rowcolor{gray!20}\multicolumn{9}{c}{\it{\textbf{MA-news}}} \\
    \hline
    1& Original         & 22.16 & 21.67 & 5.96  & 13.99 & 52.80 & 47.14 & 49.50 \\
    2& Selective Context         & 22.33 & 20.44 & 4.83  & 12.77 & 54.32 & 46.06 & 49.68 \\
    3& \textbf{SARESG}                & \textbf{24.18} & \textbf{22.95} & \textbf{6.34} & \textbf{14.76} & \textbf{55.26} & \textbf{47.97} & \textbf{51.16} \\
    \hline
    5& Truncated\_ICL    &\textbf{30.78} & 28.32 & 9.81  & 18.44 & \textbf{61.14} & 51.67 & 55.85 \\
    6& \textbf{SARESG\_ICL}          & 30.41 & \textbf{29.69} & \textbf{10.19} & \textbf{19.39} & 60.45 & \textbf{53.07} & \textbf{56.36} \\
    \hline
    \rowcolor{gray!20}\multicolumn{9}{c}{\it{\textbf{OAsum}}} \\
    \hline
    1& Original    & 24.34 & 21.52 & 6.54  & 14.61 & 58.34 & 48.30 & 52.32 \\
    2& Selective Context   & 23.66 & 19.76 & 5.36  & 13.31 & 59.19 & 47.50 & 52.29 \\
    3& \textbf{SARESG}                & \textbf{25.38} & \textbf{22.83} & \textbf{6.92} & \textbf{15.56} & \textbf{60.38} & \textbf{49.66} & \textbf{54.04} \\
    \hline
    5& Truncated\_ICL    & \textbf{24.18} & 20.81 & \textbf{6.80} & 13.84 & \textbf{59.19} & 44.55 & 50.43 \\
    6& \textbf{SARESG\_ICL}   & 23.72 & \textbf{21.00} & 6.51  & \textbf{13.94} & 58.48 & \textbf{44.98} & 50.43 \\
    \hline
    \rowcolor{gray!20}\multicolumn{9}{c}{\it{\textbf{USB}}} \\
    \hline
    1& Original     & 26.74 & 19.22 & 6.73  & 13.60  & 61.10  & 45.83 & 51.81 \\
    2& Selective Context   & 23.75 & 17.05 & 4.47  & 11.63 & 60.60  & 45.54 & 51.64 \\
    3& \textbf{SARESG}             & \textbf{29.87} & \textbf{23.20} & \textbf{8.45} & \textbf{16.29} & \textbf{64.64} & \textbf{49.79} & \textbf{55.81} \\
    \hline
    5& Truncated\_ICL       & 30.95 & 24.81 & 10.32 & 17.84 & 63.82 & 48.61 & 54.70 \\
    6& \textbf{SARESG\_ICL}         & \textbf{32.65} & \textbf{26.73} & \textbf{11.04} & \textbf{19.10} & \textbf{65.46} & \textbf{50.78} & \textbf{56.73} \\
 \bottomrule[1.5pt]
\end{tabular}
}
\caption{\textbf{The model's performance with Llama3-70b across three datasets shows that our method still outperforms most metrics.}}
\label{tab:Llama70b}
\end{table*}

\begin{table}[!t]
  \centering
    \begin{tabular}{cccc}
    \toprule
          & R1    & R2    & RL \\
    \midrule
    Sentence Retrival & 22.37 & 6.53  & 15.06 \\
    Chunk Retrival & \textbf{22.83} & \textbf{6.92} & \textbf{15.56} \\
    \bottomrule
    \end{tabular}%
  \caption{\textbf{Result comparison for retrieval by sentence and retrieval by chunk for dataset OAsum and generated with Llama3-70b.}}
  \label{tab:sentence_retrieval}%
\end{table}%

\subsection{Influence of sentence retrieval}
\label{subsec:influence_chunk_size_sentence_retrieval}

We first evaluated the performance of our retrieval system by removing the chunk-based approach and instead retrieving text at the sentence level, using the OAsum dataset. Summaries were generated using the Llama3-70b model. As shown in Table~\ref{tab:sentence_retrieval}, sentence-level retrieval yielded ROUGE-1 (R1), ROUGE-2 (R2), and ROUGE-L (RL) scores of 22.37, 6.53, and 15.06, respectively. While this approach outperformed the direct prompt method, it performed slightly worse than the chunk-based retrieval method. This discrepancy is likely because LLMs require more context to form a comprehensive understanding, and the aspects mentioned in the sentences were often difficult to identify at the sentence level. For example, if the aspect pertains to a certain man's final years, it would be a challenging task for the retrieval model to determine whether individual sentences are relevant to this aspect.

\subsection{Influence of chunk size}

To further analyze the impact of chunk size on model performance, we conducted additional experiments using the Llama3-70b model with the Ma-news dataset. Our results indicate that when the chunk size exceeds 256 words, the model is only able to retrieve two or three chunks of text. This limitation causes the retrieved content to be overly concentrated in a specific section of the text, reducing the likelihood of capturing diverse details about other aspects. Conversely, smaller chunk sizes result in overly fragmented retrieval, making it difficult for the model to identify patterns or maintain coherence.

To visualize this relationship, we created a chart illustrating the ROUGE scores for various chunk sizes. As shown in the figure, there is a clear performance gap between smaller chunk sizes and the optimal chunk size of 256 words. With a chunk size of 256, the model achieves more aspect-aligned retrieval, enabling it to generate coherent and relevant summaries. On the other hand, smaller chunk sizes lead to fragmented retrieval, which hampers the embedding model's ability to capture meaningful relationships across the text.

This analysis highlights the trade-offs between chunk size and information granularity. Larger chunks enhance coherence and relevance but may overlook broader aspects, whereas smaller chunks increase coverage at the expense of cohesion. By identifying the optimal chunk size, we aim to strike a balance between these trade-offs, ultimately improving the model’s performance in text retrieval and summarization tasks.

\begin{figure}[!t]
    \centering
    \includegraphics[width=0.8\linewidth]{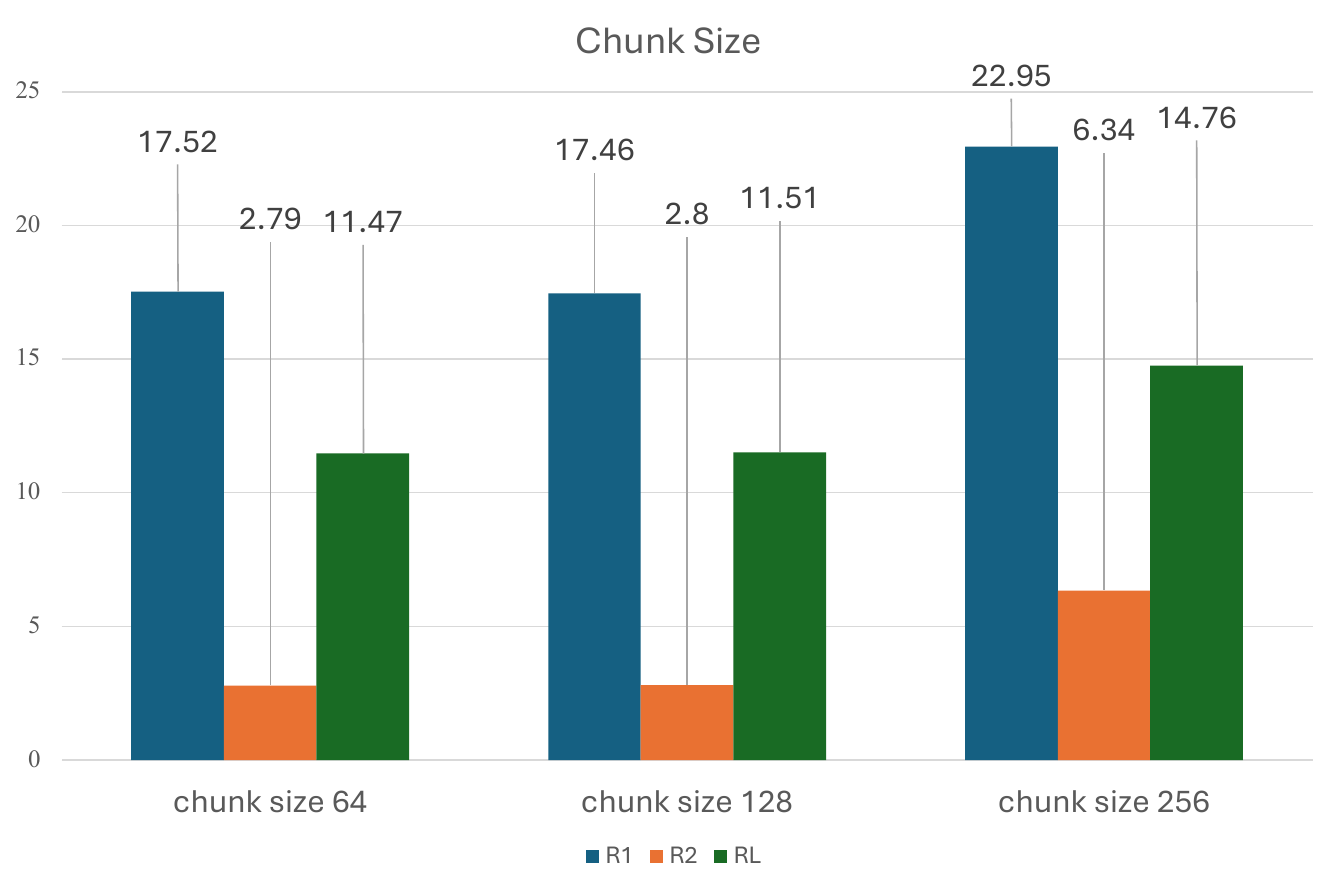}
    \caption{\textbf{Experiment results for three different chunk sizes on MA-news dataset. Results generated with Llama3-70b.}}
    \label{fig:chunk_size}
\end{figure}

\subsection{ICL sample}
As observed in the results presented in the Table~\ref{tab:mistral8*7b}, ICL generally outperformed zero-shot approaches in most cases. However, for the OAsum dataset, we observed an opposite trend: the ICL scenarios, including both truncated and SARESG approaches, achieved lower scores compared to the zero-shot cases. Upon further investigation, we identified the primary reason for this discrepancy. The randomly chosen sample used for the ICL experiments was a special case where the entire sentence consisted of only a single word followed by a colon(":"), with no sentences related to any specific aspect included in the sample. This peculiar structure likely disrupted the model's ability to effectively utilize the provided context, leading to the observed performance drop. To address this issue and ensure that ICL is also effective for the OAsum dataset, we conducted additional experiments using different samples. These new experiments were designed to include more representative and diverse examples, allowing us to evaluate whether the ICL paradigm could generalize effectively across varying data scenarios. The results of these additional experiments provided a more comprehensive understanding of the applicability and limitations of ICL in this specific dataset, thereby strengthening the robustness of our analysis.

To further investigate the impact of sample selection on ICL performance, we conducted additional research to determine whether using a specific sample related to a particular aspect would lead to improved results. For this analysis, we designed an experiment using ICL samples drawn from six different aspects, tested with the Llama3-8b model on the Ma-news dataset. The results of this experiment are visualized in the radar charts (Figure~\ref{fig:radars}).

The findings reveal that the aspect of the selected sample does influence the model's performance to some extent. However, this effect is not the primary determining factor in overall performance. While certain aspects provide slightly better results, the observed variations suggest that other factors, such as the inherent structure and relevance of the input sample, likely play a more significant role. These results underline the complexity of ICL and highlight the need for careful sample selection to optimize model outcomes.

This analysis provides valuable insights into the nuanced relationship between sample aspects and performance in ICL scenarios, paving the way for future research to systematically explore and quantify these dynamics. Further studies could examine larger-scale datasets or develop methods to identify optimal samples for diverse tasks, enhancing the generalizability and robustness of ICL approaches.

\begin{figure*}
\centering
\begin{subfigure}{0.3\textwidth}
    \includegraphics[width=\textwidth]{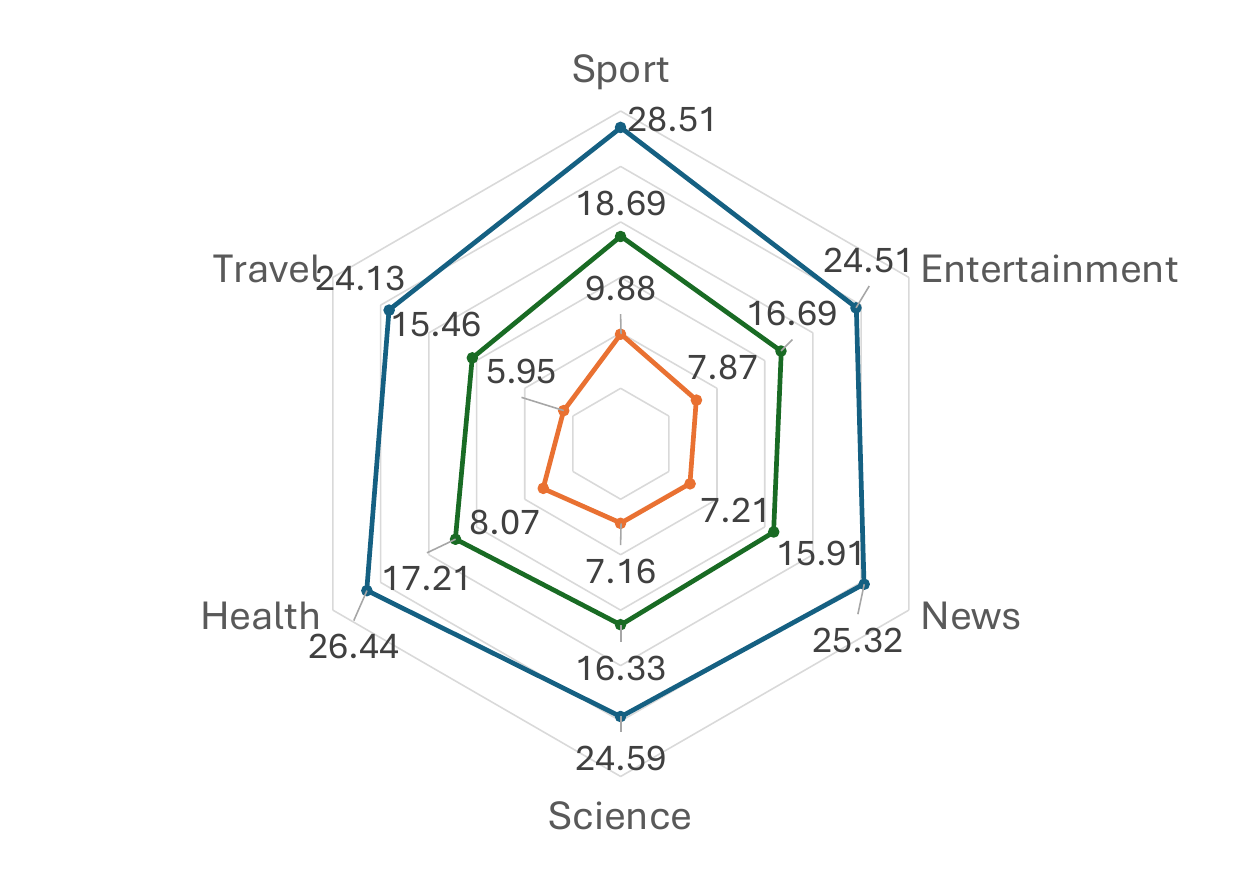}
    \caption{Health}
    \label{fig:first}
\end{subfigure}
\hspace{0.01\textwidth}
\begin{subfigure}{0.3\textwidth}
    \includegraphics[width=\textwidth]{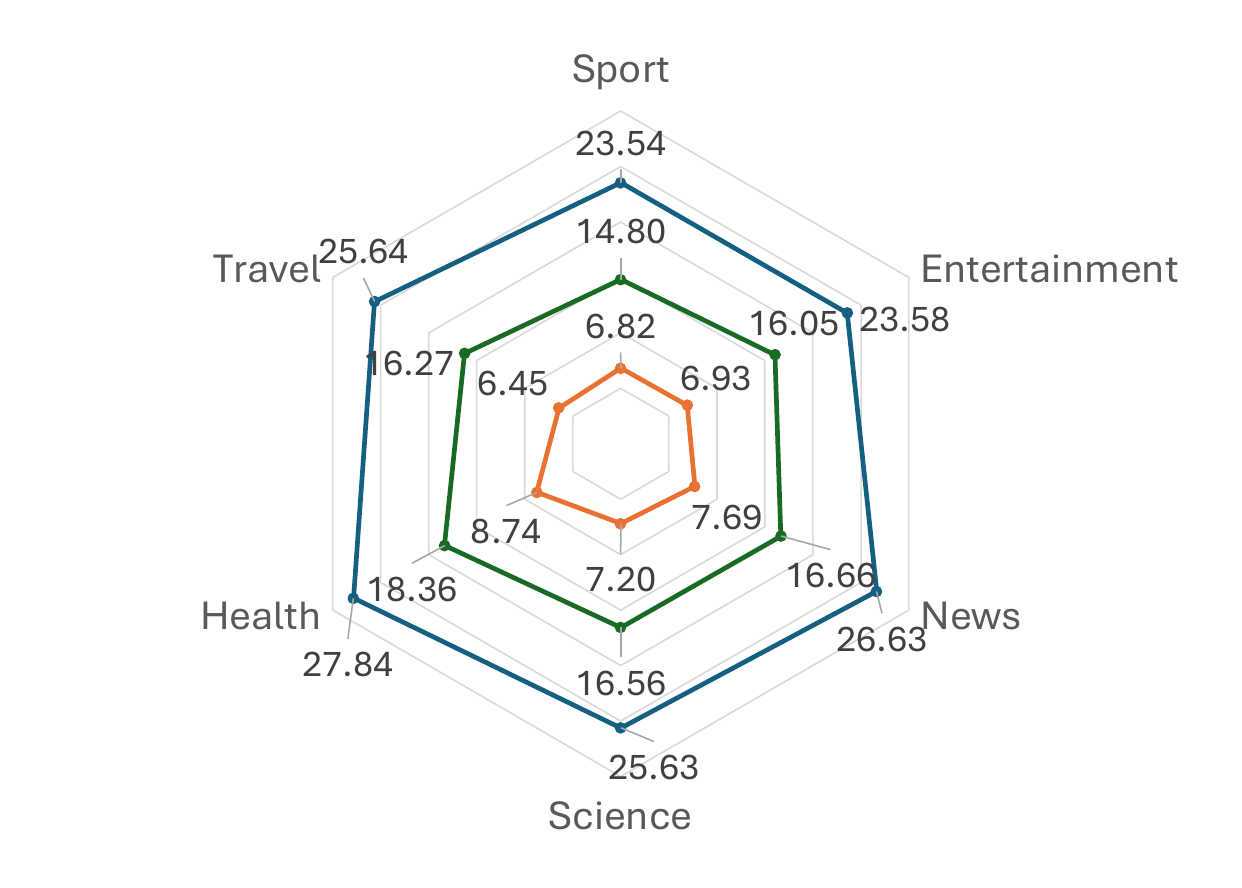}
    \caption{News}
    \label{fig:second}
\end{subfigure}
\hspace{0.01\textwidth}
\begin{subfigure}{0.3\textwidth}
    \includegraphics[width=\textwidth]{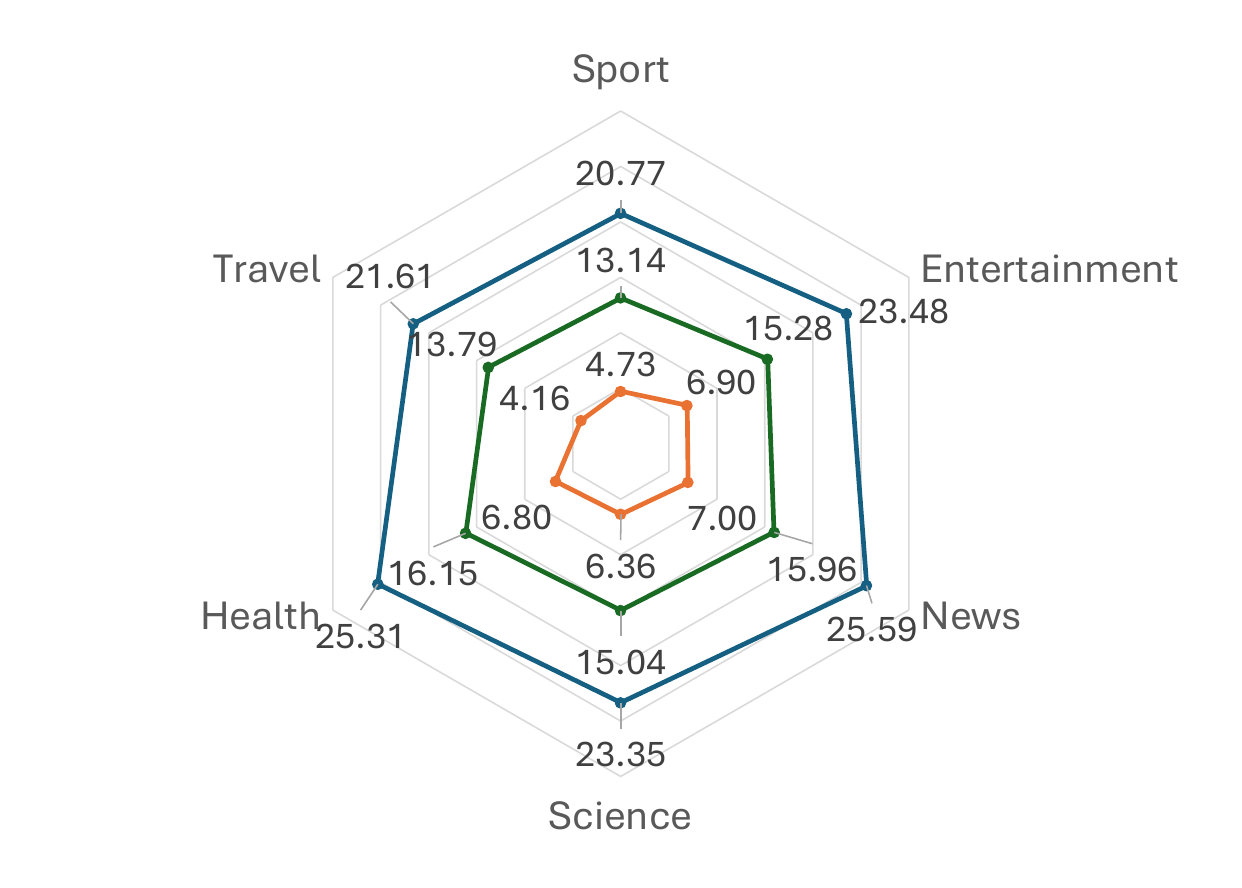}
    \caption{Science}
    \label{fig:third}
\end{subfigure}
\vskip\baselineskip
\begin{subfigure}{0.3\textwidth}
    \includegraphics[width=\textwidth]{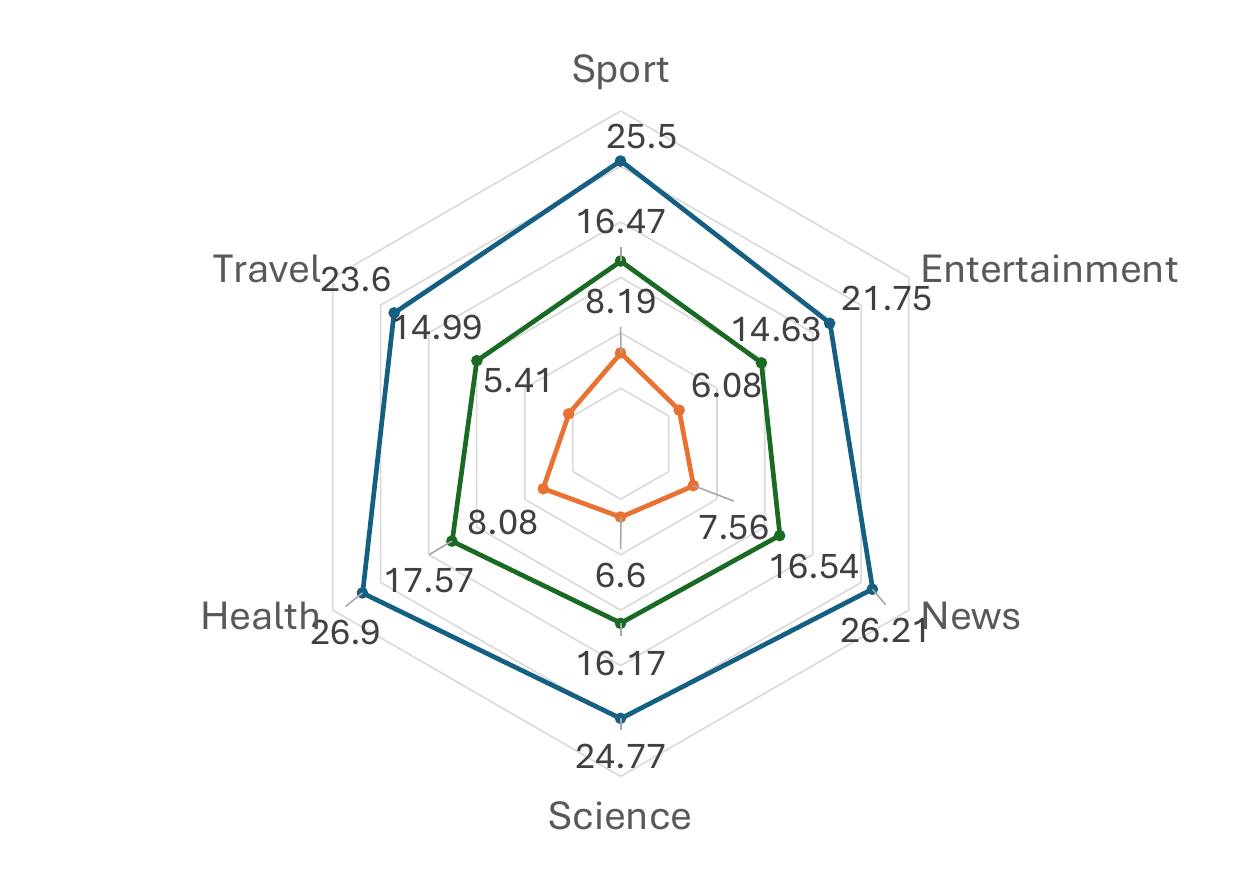}
    \caption{Sport}
    \label{fig:fourth}
\end{subfigure}
\hspace{0.01\textwidth}
\begin{subfigure}{0.3\textwidth}
    \includegraphics[width=\textwidth]{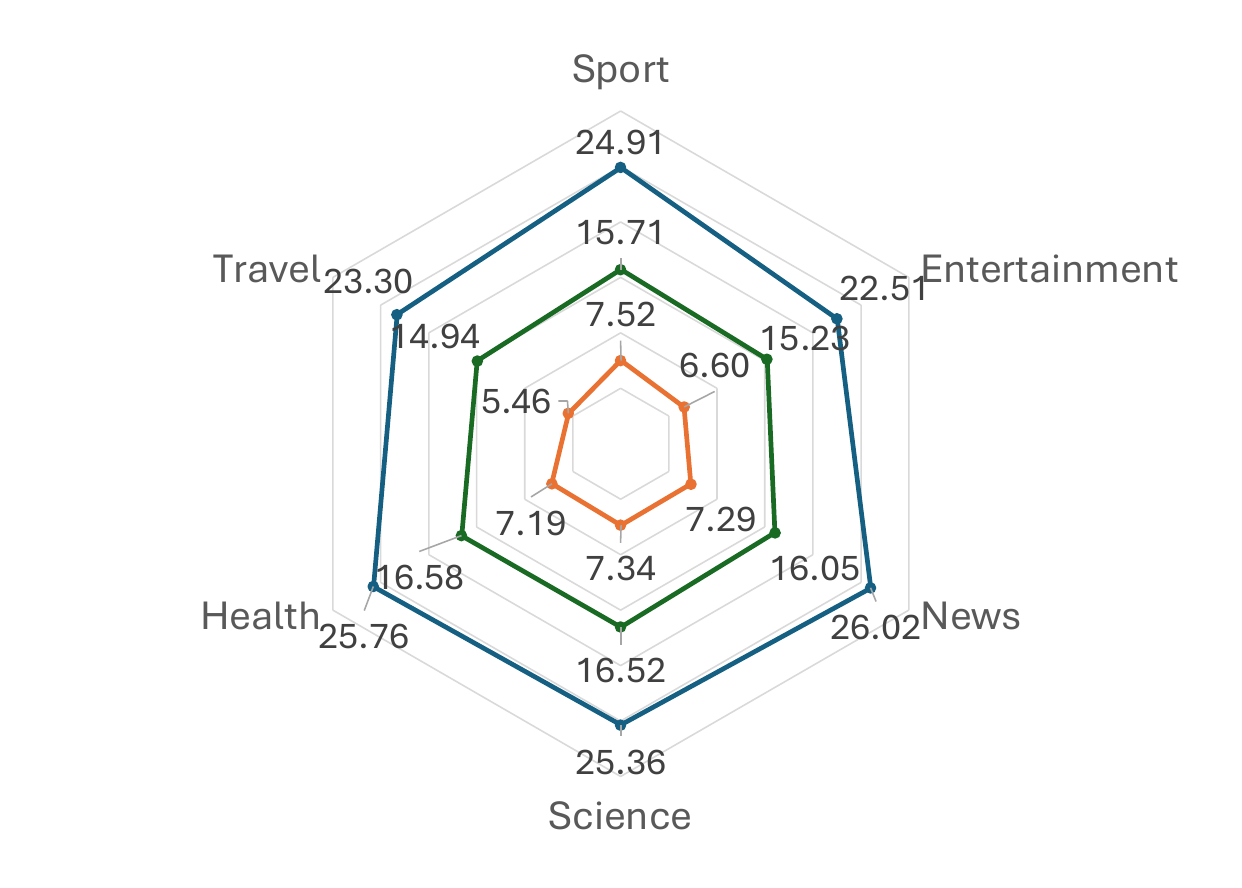}
    \caption{Travel}
    \label{fig:fifth}
\end{subfigure}
\hspace{0.01\textwidth}
\begin{subfigure}{0.3\textwidth}
    \includegraphics[width=\textwidth]{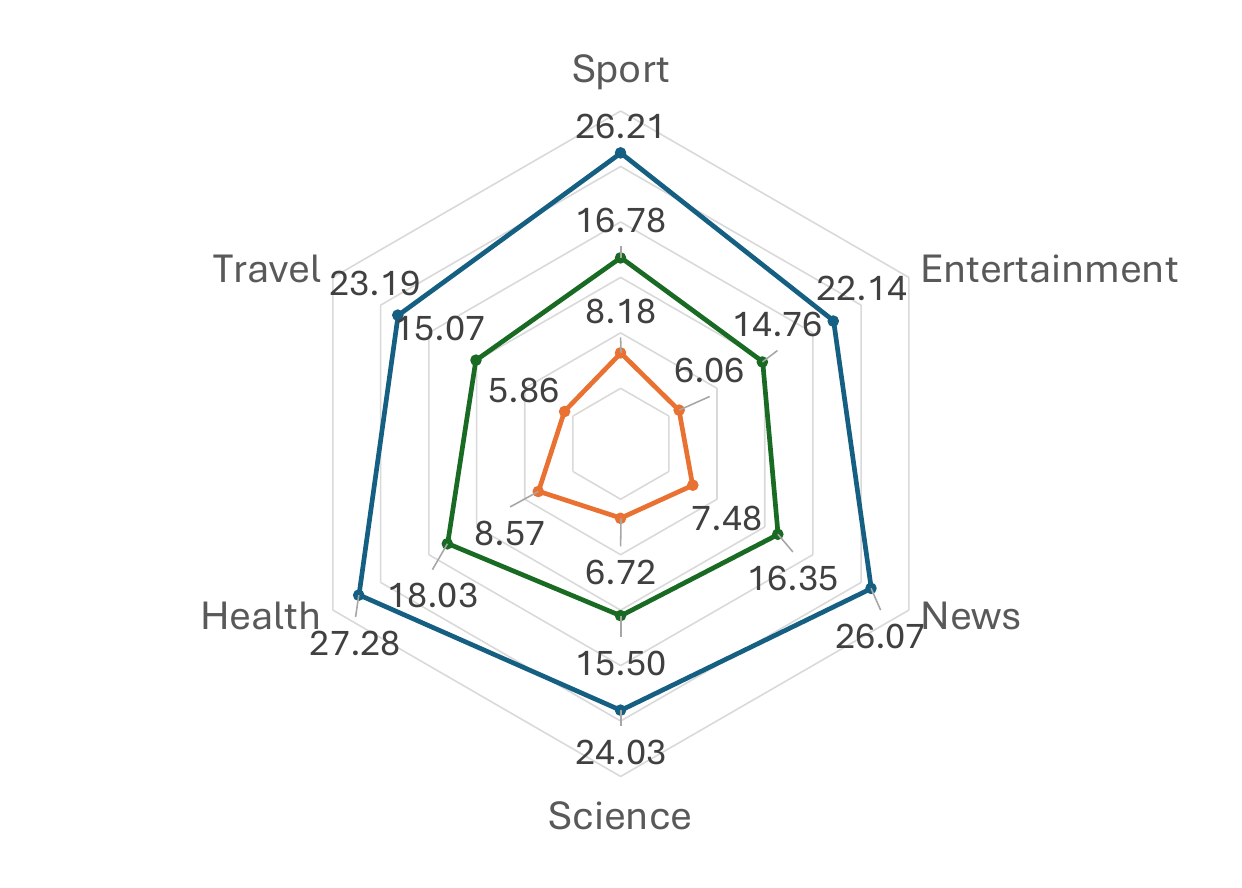}
    \caption{Entertainment}
    \label{fig:sixth}
\end{subfigure}

\caption{\textbf{Radar charts were created for all ICL samples tested on the MA-news dataset, with results generated using the Llama3-8b model. Blue is representing R1, orange is representing R2, and green is representing Rl. The captions represent the aspects of the samples, with six points on each chart corresponding to the aspects of the testing set.}}
\label{fig:radars}
\end{figure*}

\section{Discussion and Conclusion}

In this paper, we propose Self-Aspect Retrieval Summary Generation, a novel framework for aspect-based summarization.
Through the new retrieval mechanism, we can steer the language model away from unrelated information and save token space for ICL. This enables our framework to generate more aspect aligned summaries.
As the table above shows, most of the ICL were able to have a significantly higher score than the zero-shot cases, especially for smaller models, with ICL, smaller 8b model will achieve similar result with larger model like Llama3-70b with zero-shot. This might due to the reason that the larger models carries better performance and already able to generate a more precise answer, what we noticed from the smaller model is that it will generate some sentences repeatedly. While a sample was offered, the model will be more careful with the sentence they generated and have a clearer guidance.

Yet in the Table ~\ref{tab:Llama70b}, OAsum remains as the only exception here. This is because the randomly chosen samples for the OAsum are rather special.
It was a sample without any related sentences in the document. As the Figure~\ref{fig:radars}, with an alternative sample, the ICL case will still perform better than the zero shot cases with Llama3-70b model.

\section*{Limitations}
Despite all efforts, there are still a few more constraints to the method.
The first is that the ICL is highly unstable. second is that the best pruning parameters will be varied as the length of articles varied. third is that the retrieval model takes up GPU spaces for computing the similarities. And a better model requires more resources.

\bibliographystyle{IEEEtran}
\bibliography{reference}

\end{document}